\pdfoutput=1

\documentclass[11pt]{article}

\usepackage[]{acl}

\usepackage{times}
\usepackage{latexsym}
\usepackage{lineno,hyperref,graphicx}
\usepackage{booktabs, multirow}

\usepackage[T1]{fontenc}

\usepackage[utf8]{inputenc}

\usepackage{microtype}

\title{Towards Using Diachronic Distributed Word Representations as Models of Lexical Development}

\author{Arijit Gupta \\
  Cognitive Neuroscience Lab \\
  BITS Pilani, Goa Campus \\
  \texttt{\scriptsize f20180856@goa.bits-pilani.ac.in} \\\And
  Rajaswa Patil \\
  Cognitive Neuroscience Lab \\
  BITS Pilani, Goa Campus \\
  \texttt{\scriptsize f20170334@goa.bits-pilani.ac.in} \\\And
  Veeky Baths \\
  Cognitive Neuroscience Lab \\
  BITS Pilani, Goa Campus \\
  \texttt{\scriptsize veeky@goa.bits-pilani.ac.in}}

\begin{document}
\maketitle
\begin{abstract}
Recent work has shown that distributed word representations can encode abstract information from child-directed speech. In this paper, we use diachronic distributed word representations to perform temporal modeling and analysis of lexical development in children. Unlike all previous work, we use temporally sliced corpus to learn distributed word representations of child-speech and child-directed speech under a curriculum-learning setting. In our experiments, we perform a lexical categorization task to plot the semantic and syntactic knowledge acquisition trajectories in children. Next, we perform linear mixed-effects modeling over the diachronic representational changes to study the role of input word frequencies in the rate of word acquisition in children. We also perform a fine-grained analysis of lexical knowledge transfer from adults to children using Representational Similarity Analysis. Finally, we perform a qualitative analysis of the diachronic representations from our model, which reveals the grounding and word associations in the mental lexicon of children. Our experiments demonstrate the ease of usage and effectiveness of diachronic distributed word representations in modeling lexical development.
\end{abstract}

\section{Introduction}
\label{sec-intro}

Human-like linguistic generalization plays a key role in developing better models for natural language processing \cite{linzen-2020-accelerate}. Modeling the lexical development in children is an important aspect of demystifying the dynamics of human language learning. Lexical development in children is a holistic and complex phenomenon involving noisy multimodal interactions and underlying various psycholinguistic processes. Previous research in child language acquisition has shown that infants are capable of lexical processing of words through their semantic and syntactic distributional structures \citep{doi:10.1177/0956797609358570, doi:10.1080/15475440903507905}. Recently, the paradigm of word embeddings from deep-learning-based computational semantics has pushed the frontiers in modeling such distributional structures of words \citep{10.5555/2999792.2999959, wang2020survey}. Consequently, word embeddings have been used to study various aspects of child-speech and child-directed adult speech \citep{huebner_willits_2018, fourtassi-etal-2019-development, fourtassi-2020-word}.

Recent advances in computational modeling for distributional semantics have made it possible to study the diachronic semantic shifts in a given corpus \citep{kutuzov-etal-2018-diachronic}. Using temporally-wide large-scale corpora, diachronic word embeddings can be used to study the underlying linguistic and non-linguistic dynamics of change and development in human language \citep{kutuzov-etal-2018-diachronic}. Given the availability of such corpora for child-speech and child-directed adult speech \citep{MacWhinney2000}, a similar framework can be designed to model the lexical development in children.

This paper explores the usability of diachronic distributed word representations\footnote{the terms ``distributed word representations'', ``word vectors'', and ``word embeddings'' have been used interchangeably in this paper.} in cognitive modeling and analysis of the lexical development in children. Unlike previous work, we use temporally sliced data to learn distributed word representations of child-speech and child-directed speech under a curriculum-learning-like setting. Through our experiments we show that diachronic word representations can be very effective in capturing various empirical and qualitative aspects of lexical development in children.

\begin{figure*}[thbp]
    \centering
    \includegraphics[width=0.75\textwidth]{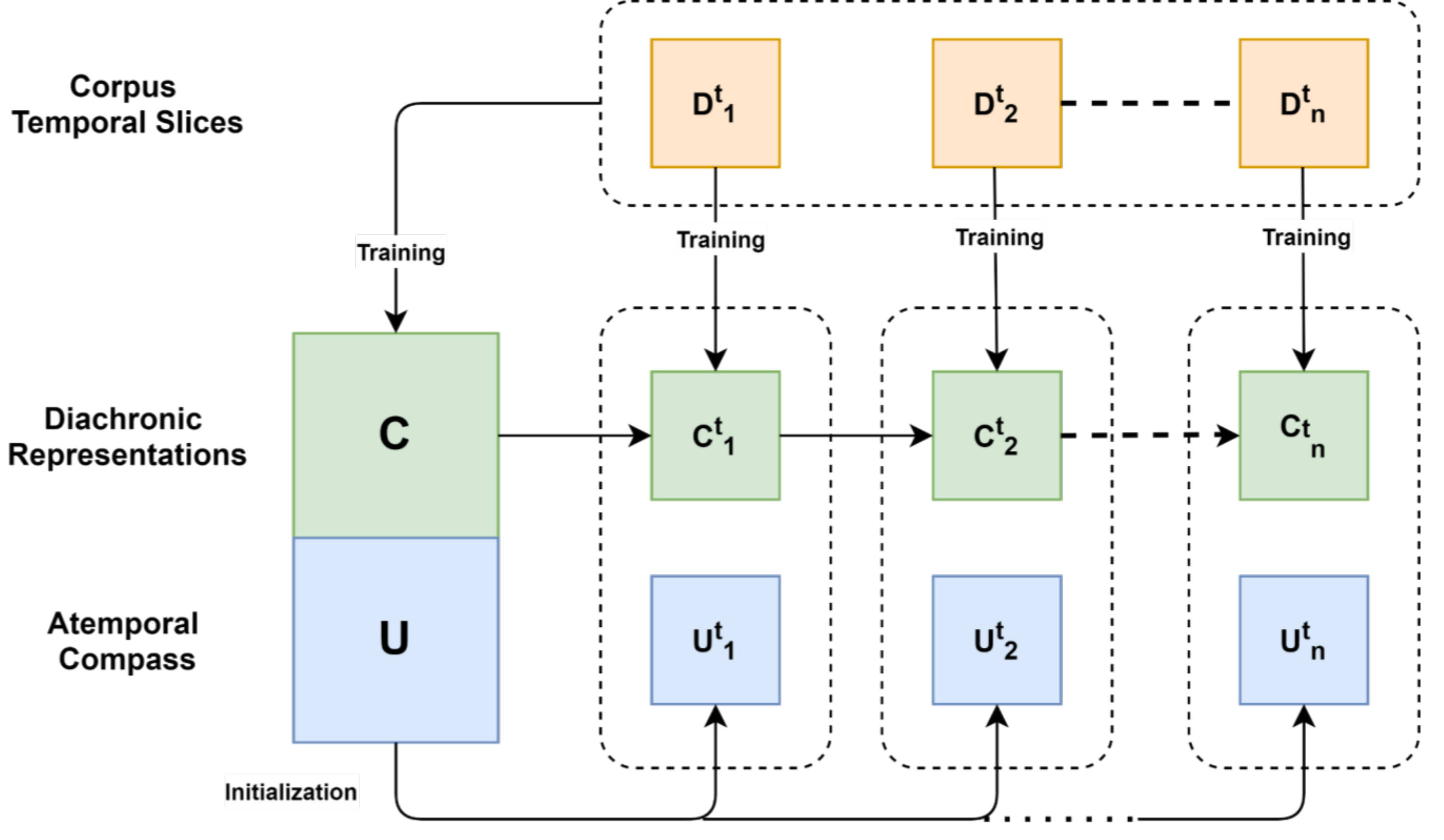}
    \caption{The architecture for the incremental diachronic compass-based word embedding model. In order to  temporally align the word embeddings from each of the corpus temporal slices, one of  the layers $(U_i^t)$ is initialized from the atemporal compass and frozen during the temporal fine-tuning. The representations for each time-step $C_i^t$ are initialized with the fine-tuned representations for the previous time-step $C_{i-1}^t$, and fine-tuned on its corresponding corpus temporal slice $D_i^t$ (Section~\ref{sec-model}).}
    \label{fig:model-architecture}
\end{figure*}

\section{Background}
\label{sec-background}

The meanings of words change over time, owing to a variety of linguistic and non-linguistic factors. This phenomenon has been termed as semantic shifts \citep{bloomfield_1933} in historical linguistics. Due to an increasing availability of large corpora, several data-driven methods have been proposed to study such semantic shifts. Words can be represented as continuous vectors \citep{rumelhart_mcclelland_1986, ELMAN1990179}. As the meaning and context of a word changes, it’s vector changes accordingly \citep{kutuzov-etal-2018-diachronic}. Recent work has shown that semantic shifts are not always closely related to the changes in word frequencies \citep{kutuzov-etal-2018-diachronic}. Previous works have also shown that distributed word representations \citep{10.5555/1861751.1861756, baroni-etal-2014-dont} outperform frequency based methods in detecting semantic shifts \citep{kulkarni2014statistically}. These models use deep-learning based word-embedding vectors \citep{mikolov_chen_corrado_dean_2013}, produced from word co-occurrence relationships.

Semantic shifts were first analyzed on year-wise data by \citet{kim-etal-2014-temporal} using a prediction-based word embedding model. \citet{kim-etal-2014-temporal} used a distributional continuous skip-gram model with negative sampling \citep{mikolov_chen_corrado_dean_2013}, which was later proven to be superior in semantic shift analysis as compared to PPMI-based distributional models \citep{hamilton-etal-2016-diachronic}. Most modern word embedding models are inherently stochastic \citep{kutuzov-etal-2018-diachronic}, and produce word representations in different vector spaces on each run. To overcome this, the models need to be aligned to one common vector space. This can be done by performing certain linear transformations on the diachronic word embeddings \cite{kulkarni2014statistically, zhang-etal-2015-omnia}. More recently, \citet{carlo_bianchi_palmonari} proposed a relatively simple temporal compass based alignment, which showed significant improvements over the previous approaches.

Diachronic embeddings have been used across a variety of applications, ranging from linguistic studies to cultural studies. They have been successfully used for tasks like event detection, tracing popularity of entities \citep{kutuzov-etal-2018-diachronic}, and identifying scientific trends \cite{sharma-etal-2021-drift}. Even though there have been a significant number of studies which use static distributed word representations for analyzing child-directed speech \citep{huebner_willits_2018_another, huebner_willits_2018, fourtassi-etal-2019-development, fourtassi-2020-word}, the usability of diachronic embeddings as models of lexical development has not been explored yet.\footnote{A parallel study by \citet{jiang2020exploring} focuses on using diachronic word embeddings to study the child-directed speech. Whereas, in this work we focus on directly modeling the lexical development in children.} The closest work is presented by \citet{huebner_willits_2018_another}, where they train sequential deep-learning models on age-ordered child-directed speech data. In this study, we use diachronic word embeddings trained on child-speech data,\footnote{This has not been explored before to the best of our knowledge} to construct a temporal representational model for the mental lexicon in children.

\section{Modeling}
\label{sec-methodology}
Given a corpus of child-speech, a diachronic word-embedding model can be trained over its temporal slices. The distributed-latent representations from the model can then be probed for lexical knowledge at any given point of time. Consequently, the \emph{lexical development} can be simply captured by comparing these distributed representations over some interval of time. In this section, we describe our cognitively motivated diachronic modeling method (Section~\ref{sec-model}), and the required pre-processing of the child-speech corpus (Section~\ref{sec-data}). We discuss the usability of the trained diachronic distributed representations for modeling lexical development in Section~\ref{sec-analysis}.

\subsection{Data}
\label{sec-data}

Similar to all the previous works, we use the CHILDES corpus \citep{MacWhinney2000} for our experiments. The corpus consists of speech-transcripts of first language acquisition by children. It contains transcriptions in $26$ languages, spanning across $130$ corpora of children interacting in different environments, including spontaneous interactions, as well as controlled classroom learning. For the current study, we use the data from the American-English speaking children. Due to an imbalance in the data distribution, we discard the data beyond the first three years of age.

Unlike child-directed adult speech, child speech is highly noisy in the early months. It is only after the age of $18$ months, that children start combining two words or single-word phrases in situations in which they both are relevant and having roughly equivalent status \citep{bavin_naigles_2017}. Hence, we consider the data after the age of $18$ months only. This results in a corpus that is temporally spread across $19$ months (age=$18$ months to age=$36$ months). It contains $2798$ speech transcripts; $1,321,772$ word tokens; $13,812$ word types; and $405,596$ utterances, collected from $28$ different studies involving $188$ children ($99$F and $89$M) and their guardians.

We tokenize the corpus with whitespace delimitation. We remove all the punctuation from the corpus as they do not contribute lexically in any way. Further, all the proper nouns are replaced with a generalized token: $[NAME]$. We set the temporal granularity of our diachronic models to a month. Hence, we split the corpus into $19$ month-wise temporal slices.

\subsection{Model}
\label{sec-model}

A diachronic word embedding model usually comprises two major components: a base word-embedding model, and a mechanism to align the representations across different temporal data slices. We use $word2vec$ as our base word-embedding model (skip-gram with negative sampling - SGNS variant) \citep{mikolov_chen_corrado_dean_2013}. For aligning the word-embeddings, we employ a slightly modified version of the compass-based alignment method proposed by \citet{carlo_bianchi_palmonari}.

$word2vec$ is a shallow, two-layered neural-network word-embedding model. It takes a large corpus of words as input, and generates a multi-dimensional distributed vector space, with each word being assigned a vector. The $word2vec-SGNS$ model in particular takes a one-hot encoded word identity vector as input and predicts its surrounding context words. Formally, given a sequence of input words $w_{1},w_{2},w_{3}....w_{T}$, the objective of the model is to maximize the average log probability:

\begin{equation}
    \frac{1}{T}\sum_{t=1}^{T}\sum_{-c<j<c,j\neq0}logp(w_{t+j}|w_{t})
\end{equation}

where $c=5$ is the size of the training context window. The skip-gram formula defines $p(w_{t+j}|w_{t})$ using a negative sampling method \citep{mikolov_chen_corrado_dean_2013}.

Some words are quite frequent in the corpus while others are less frequent; to deal with this, a sub-sampling strategy is used. Each word in the training set $w_{i}$ is discarded with the probability:

\begin{equation}
    P(w_{i}) = 1-\sqrt{\frac{t}{f(w_{i})}} 
\end{equation}

where $f(w_{i})$ is the frequency of the word and $t$ is some selected threshold. This accelerates the learning process and improves the accuracy of the vector representations of rare words. We use a word representation dimension of $d=100$ for the $word2vec-SGNS$ model.

To work with diachronic word embeddings, the embedding spaces generated by the models trained at each time-step need to be aligned to a common embedding space. This ensures that the embeddings across different time-steps can be compared directly. We implement a temporal compass-based model to align the embedding spaces \citep{carlo_bianchi_palmonari}. As the $word2vec-SGNS$ model is a two-layered neural network, the context embeddings are encoded in the output layer parameters $(U)$ and target word embeddings are encoded in the input layer parameters $(C)$ (Figure~\ref{fig:model-architecture}). Given a corpus $D$, divided into $n$ temporal slices: $D_1^t, D_2^t, ...D_n^t$; the model $(C,U)$ is first trained on the entire corpus $D={D_1^t, D_2^t, .... D_n^t}$ as shown in Figure~\ref{fig:model-architecture}. The input layer parameters $(C)$ are used as the distributed representations in our experiments. Whereas, the output layer parameters $(U)$ are used as an atemporal compass to align these distributed representations. Hence, the output layer parameters are frozen and unchanged for further training, such that: $U=U_1^t=U_2^t=U_3^t=... U_n^t$. The diachronic representations for each temporal slice in the corpus $(D_i^t)$ are then obtained by fine-tuning the input layer parameters $(C_i^t)$ on its corresponding temporal slice $(D_i^t)$.

The fine-tuning for each temporal slice $(D_i^t)$ is resumed with the representations from the previous temporal slice $(C_{i-1}^t)$. This is done by initializing the input parameters $(C_i^t)$ with the already fine-tuned input parameters from the previous temporal slice $(C_{i-1}^t)$. Where the parameters for the newly acquired words in $(D_i^t)$ are initialized randomly. This ensures that the diachronic model captures the lexical development in a cognitively plausible incremental way, following the paradigm of curriculum-learning in children. Formally, given a slice $D^{t}$ the training procedure for an input $(w_{k},\gamma(w_{k}))$ is defined as the following optimization problem:

\begin{equation}
    \resizebox{0.8\columnwidth}{!}{$max\Big($log$P(w_{k}|\gamma(w_{k}))\Big) = \sigma\Big(\vec{u}_{k}\cdot\vec{c}^{\hspace{0.005\textwidth}t}_{\gamma(w_{k})}\Big)$}
\end{equation}

carried out on $C^{t}$ where the function $\sigma$ is calculated using negative sampling, $\gamma(w_{k}) = (w_{1},w_{2},...w_{M})$ is the set of M words that appear in the context of $w_{k}$ ($\frac{M}{2}$ being the size of the window), $\vec{u}_{k} \in U$ is the atemporal target embedding of the word $w_{k}$ and $\vec{c}^{\hspace{0.005\textwidth}t}_{\gamma(w_{k})}$ is the mean of the temporal context embeddings.

We train separate models for child-speech and child-directed speech, each providing distributed word representations independent of each other. We use these models to compare the lexical development in child-speech and child-directed speech. We also train both these models in both, the proposed incremental manner, and in a non-incremental way \citep{carlo_bianchi_palmonari} for ablation purposes. We train the models with three different random initialization seed values, and report the results averaged across the random seeds.

\begin{figure*}[htb]
    \centering
    \includegraphics[width=\textwidth]{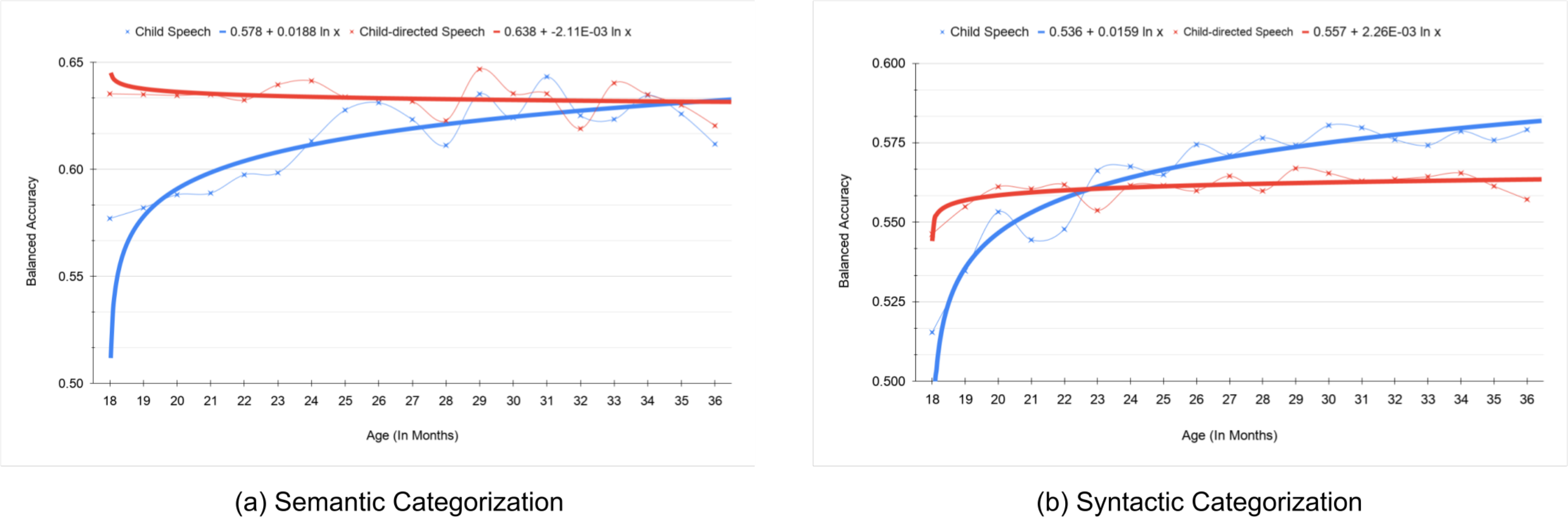}
    \caption{Month-wise Balanced Accuracy scores for the lexical categorization experiment with the non-incremental diachronic representations for the child-speech and child-directed adult speech.}
    \label{fig:BA-non-incremental}
\end{figure*}

\begin{figure*}[htb]
    \centering
    \includegraphics[width=\textwidth]{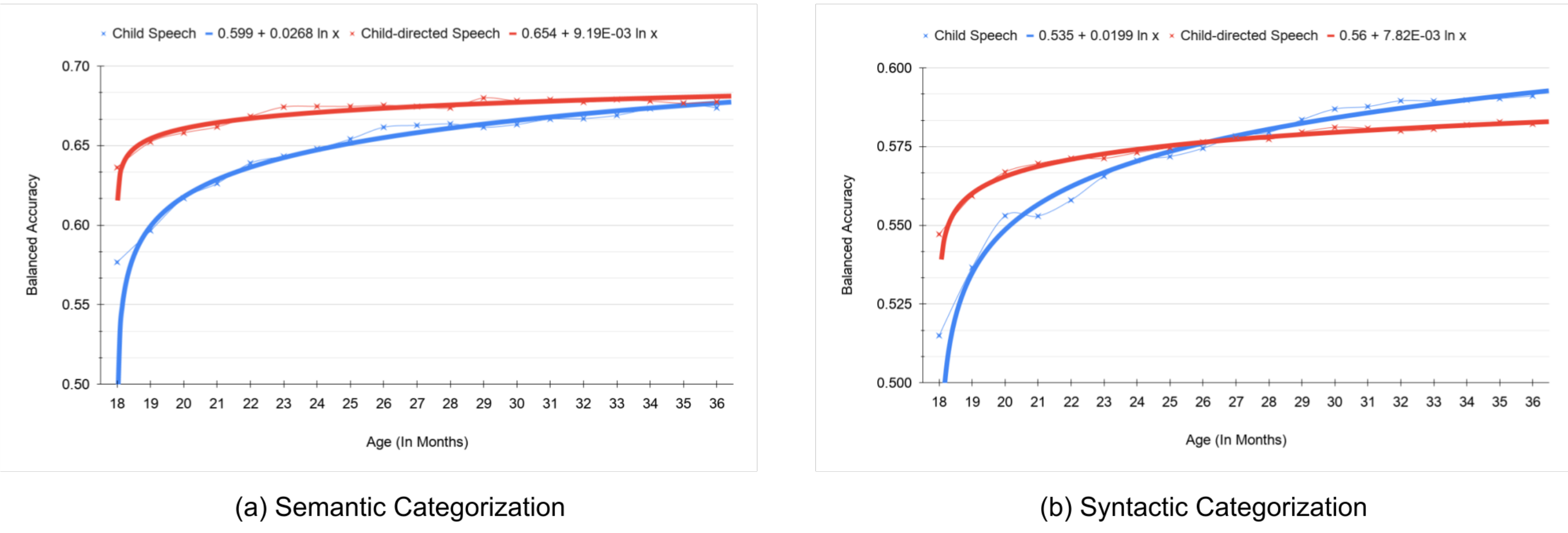}
    \caption{Month-wise Balanced Accuracy scores for the lexical categorization experiment with the incremental diachronic representations for the child-speech and child-directed adult speech.}
    \label{fig:BA-incremental}
\end{figure*}

\begin{table}[thbp]
\centering
\resizebox{\columnwidth}{!}{
\begin{tabular}{@{}|c|c|c|@{}}
\toprule
\textbf{Data}                  & \textbf{Semantic} & \textbf{Syntactic} \\ \midrule
\textbf{Child-speech}          & 141               & 347                \\
\textbf{Child-directed speech} & 184               & 597                \\
\textbf{Combined}              & \textbf{126}               & \textbf{335}                \\ \bottomrule
\end{tabular}
}
\caption{The number of common probe-words across all the $19$ months of data.}
\label{tab:probe-word-stats}
\end{table}

\section{Analysis}
\label{sec-analysis}

Given the parameterized representational nature of our model, it can be used to study the various empirical and qualitative aspects of lexical development in children. Here, we show the usability of our model by performing three experiments, and qualitative analysis\footnote{We report the details for the qualitative analysis in Appendix~\ref{sec:appendix-qualitative}.} of its representations. Similar to the previous studies, we use a set of syntactic and semantic probe words (that occur frequently in child-speech) for this purpose. These probe words are derived from the MacArthur-Bates Communicative Development Inventory (MCDI). We obtain the probe words from the data used in \citet{huebner_willits_2018_another}.

We consolidate the vocabularies and obtain the common words appearing across all of the temporal slices. From this set of common words, we only consider the ones that are a part of the previously obtained semantic and syntactic probe words, where the rest of the words are discarded for analysis. The final set of syntactic probe words is classified into eight part-of-speech categories, and the final set of semantic probe words is classified into $24$ abstract semantic categories. The statistics for the final set of probe words are given in Table~\ref{tab:probe-word-stats}.

\begin{table*}[thbp]
\centering
\begin{tabular}{@{}|c|c|c|c|c|@{}}
\toprule
\textbf{Category}   &   \textbf{Word frequency in:}   &   \textbf{$\beta_{f}$}    &   \textbf{$\beta_{t}$}  &   \textbf{$\epsilon^{(t)}_{w_{i}}$}\\
\midrule
\multirow{2}{*}{\textbf{Syntactic Probe Words}} & {Child-speech} & -0.272 & -0.047 & 1.008 \\ \cline{2-5}
& {Child-directed speech} & \textbf{-0.096} & -0.104 & 0.714\\ \cline{1-5}
\multirow{2}{*}{\textbf{Semantic Probe Words}} & {Child-speech} & -0.322 & -0.141 & 1.156\\ \cline{2-5}
& {Child-directed speech} & \textbf{-0.224} & -0.182 & 1.058\\
\bottomrule
\end{tabular}
\caption{The results for the linear mixed random-effects models fitted on semantic change values $\Delta^{(t)}w_i$ in child-speech, with respect to the word frequencies in child-speech and child-directed speech.}
\label{tab:freq-stats}
\end{table*}

\subsection{Lexical Category Learning}
\label{sec-lex-category}

Performing lexical categorization is an important aspect of lexical development. The representational nature of our model allows it to perform categorization with any vector-similarity-based measure. We borrow the task of lexical categorization from \citet{huebner_willits_2018}. As the probe words are divided into well-defined syntactic and semantic categories, we quantify the category learning ability of our model using a balanced accuracy measure \citep{huebner_willits_2018}.

We use signal detection theory to calculate the measure of interest. For each probe word, a comparison is done with the other probe words in their own category as well as the ones in different categories. We use cosine-similarity measure $cosine (w,w')$ for this purpose. Two words are classified into the same category only if $cosine (w,w') \geq r$, where $r \in (0,1)$ is a threshold value. Each correct classification is recorded as a $hit$ and the incorrect classification is recorded as a $miss$. This is finally used to calculate a balanced accuracy measure $(BA)$:

\begin{equation}
    BA = \frac{TPR + TNR}{2}
\end{equation}

where $TPR$ (true-positive rate) is the $sensitivity$ and $TNR$  (true-negative rate) is the $specificity$. For each temporal slice, the threshold value $(r)$ is calculated to maximize the balanced accuracy in an iterative way (with a step-size of $1e-3$).

We perform the lexical categorization task with both non-incremental \citep{carlo_bianchi_palmonari} and incremental modeling (ours) approaches. We calculate the balanced accuracy measure for both the syntactic and semantic probe words for each month. This gives us a trajectory of lexical category learning in child-speech as shown in Figure~\ref{fig:BA-non-incremental} and Figure~\ref{fig:BA-incremental}. We also plot the month-wise balanced accuracies for the child-directed adult speech, which gives us a trajectory of the lexical knowledge present in it. We model these trajectories by applying a temporal logarithmic fit\footnote{The final curve fit equations are given in the legends of Figure~\ref{fig:BA-non-incremental} and Figure~\ref{fig:BA-incremental}.} over the balanced accuracies values in the following manner:

\begin{equation}
    BA = \alpha + \beta \times log_{e}(t) 
\end{equation}

where $BA$ is the balanced accuracy, $\alpha$ is the intercept, and $\beta$ is the log-curve coefficient.

Overall, we observe that the lexical categorization knowledge in children increases logarithmically over time, eventually saturating around the almost-constant level of existing categorization knowledge in adults (child-directed speech). For the semantic categorization, we observe that the incremental model gives a maximum balanced accuracy\footnote{calculated with the representations from child-speech.} of $0.6738$ (t=$36$ months), and the non-incremental model gives a maximum balanced accuracy of $0.6118$ (t=$36$ months). Similarly, for the syntactic categorization the maximum balanced accuracy scores are $0.5911$ (t=$36$ months) and $0.5791$ (t=$36$ months) for the incremental and non-incremental models respectively.

Hence, the incremental model shows a significant improvement of $10.13\%$ for the semantic categorization, and $2.07\%$ for the syntactic categorization. The incremental model also shows smoother and monotonic balanced accuracy trajectories. This demonstrates the importance of the cognitively motivated curriculum learning method in our model, where the incremental model captures better lexical knowledge in its distributed representations than the non-incremental model.

\subsection{Word Frequencies and Lexical Development}
\label{sec-word-freq}

Word-frequencies have been extensively analyzed under various aspects of child-speech and child-directed speech in many previous works \citep{ambridge2015ubiquity}. Diachronic word embeddings have also been used to postulate laws mapping frequency-based measures to the historical semantic shifts of words. One such law by \citet{hamilton-etal-2016-diachronic} states that frequent words change more slowly. This can be formally expressed as:

\begin{equation}
    \Delta w_{i} \propto f(w_{i})^{\beta_{f}}
\end{equation}

where $\Delta w_{i}$ is the rate of semantic change, $f(w_{i})$ is the frequency of the word $w_{i}$ and $\beta_{f}$ is a negative power as per the relation.

Given the word-frequency data, and the distributed word representations from a diachronic model, similar effects of word-frequencies can be inspected for lexical development in children. In order to demonstrate this, we borrow a modeling approach by \citet{hamilton-etal-2016-diachronic}. Semantic change for each word at consecutive time steps $(t, t+1)$ can be calculated as:

\begin{equation}
    \Delta^{(t)} w_{i} = 1 - cosine\big(w_{i}^{(t)},w_{i}^{(t+1)}\big)
\end{equation}

In the context of child speech, this value can be looked upon as the update in the mental representation of the word in the child’s mental lexicon. The trajectory of a word’s semantic change values can then be thought of as the process of acquiring (learning and grounding) the meaning of that word.

Following \citet{hamilton-etal-2016-diachronic}’s approach, we log transform and normalize the $\Delta^{(t)}w_i$ values. The $\Delta^{(t)}w_i$ values that are less than $0.05$ are not considered in order to maintain numerical stability in the logarithm and to ignore the insignificant changes in the representations. These new values are denoted as $\bar{\Delta}^{(t)}w_{i}$. We then fit a linear mixed random-effects model in the following manner:

\begin{equation}
    \bar{\Delta}^{(t)}w_{i} = \beta_{f}log\Big(f^{(t)}(w_{i})\Big) + \beta_{t} + z_{w_{i}} + \epsilon^{(t)}_{w_{i}} 
\end{equation}

where $\beta_{f}$ and $\beta_{t}$ are fixed effects for $frequency$ and $time$ respectively, $z_{w_{i}}$ is the random intercept and $\epsilon^{(t)}_{w_{i}}$ is the $error$ term.

We use the representations from the incremental model to obtain the $\Delta^{(t)}w_i$ values for child-speech. We fit separate linear mixed random-effects models for frequency data from child-speech and child-directed speech as shown in Table~\ref{tab:freq-stats}.\footnote{all the obtained model fits are statistically significant with p-value $< 0.05$} Similar to the findings of \citet{hamilton-etal-2016-diachronic}, we find that $\beta_f$ takes negative values across both syntactic and semantic probe words, and word frequencies from both the child-speech and the child-directed speech. While it can be argued that the $word2vec$ model’s dependence on word co-occurrence statistics might superficially induce negative $\beta_f$ values for frequencies from child-speech, the negative $\beta_f$ values for frequencies from the child-directed speech are independently obtained (given that the model trained on child-speech is not exposed to the data from child-directed speech at any point of time). Hence, we majorly focus on the $\beta_f$ values from child-directed speech, which are obtained from the input word frequencies to the children. These negative $\beta_f$ values are, in general, in good agreement with all the previous studies on the role of input word frequency in word acquisition \citep{ambridge2015ubiquity}. Where it is known that higher single-word frequencies are usually associated with quicker word acquisition (which translates to smaller semantic change values $\Delta^{(t)}w_i$ with respect to the temporal slices of the word exposure).

While the $\beta_f$ values for semantic probe words are significantly negative as expected, the $\beta_f$ values for syntactic probe words, although negative, are slightly close to $0$. While this is only in a weak agreement with most of the previous studies, it is important to note that these studies inspect the role of input word frequencies with respect to specific syntactic constructs and categories \citep{ambridge2015ubiquity}. Whereas our results are representative of all the syntactic words in general.

\subsection{Representational Similarity Analysis (RSA)}
\label{sec-lex-acquisition}

Lexical acquisition in children is pragmatized by the child-directed adult speech \cite{clark2017lexical}. While our results from the Lexical Categorization task (Section~\ref{sec-lex-category}) implicitly depict the lexical knowledge transfer from adult to child, a more fine-grained analysis can be performed by directly comparing the distributed representations for child-directed speech and child-speech. Recent work in natural language processing research has focused on using Representational Similarity Analysis (RSA) \cite{rsa2000, rsa} for various interpretability studies \cite{abnar-etal-2019-blackbox, gauthier-levy-2019-linking, lepori-mccoy-2020-picking, merchant-etal-2020-happens}. 

We use RSA to compare the diachronic representational geometries of child-speech and child-directed speech. Following the settings used by \citet{lepori-mccoy-2020-picking}, we use Spearman's correlation ($\rho$) as the similarity metric ($sim$). For each time-step (i.e. month-wise), we first obtain the individual geometries for the corresponding representations for child-directed speech and child-speech by using the dissimilarity metric: $1-sim$. For a fair comparison with the results from Section~\ref{sec-lex-category}, we only use the similar set of semantic and syntactic probe words to obtain the representational geometries. The final similarity value between the representational geometries is then obtained by using the similarity metric ($sim$).\footnote{all the obtained correlation values are statistically significant with p-value $< 0.05$}

\begin{figure}[thbp]
    \centering
    \includegraphics[width=\columnwidth]{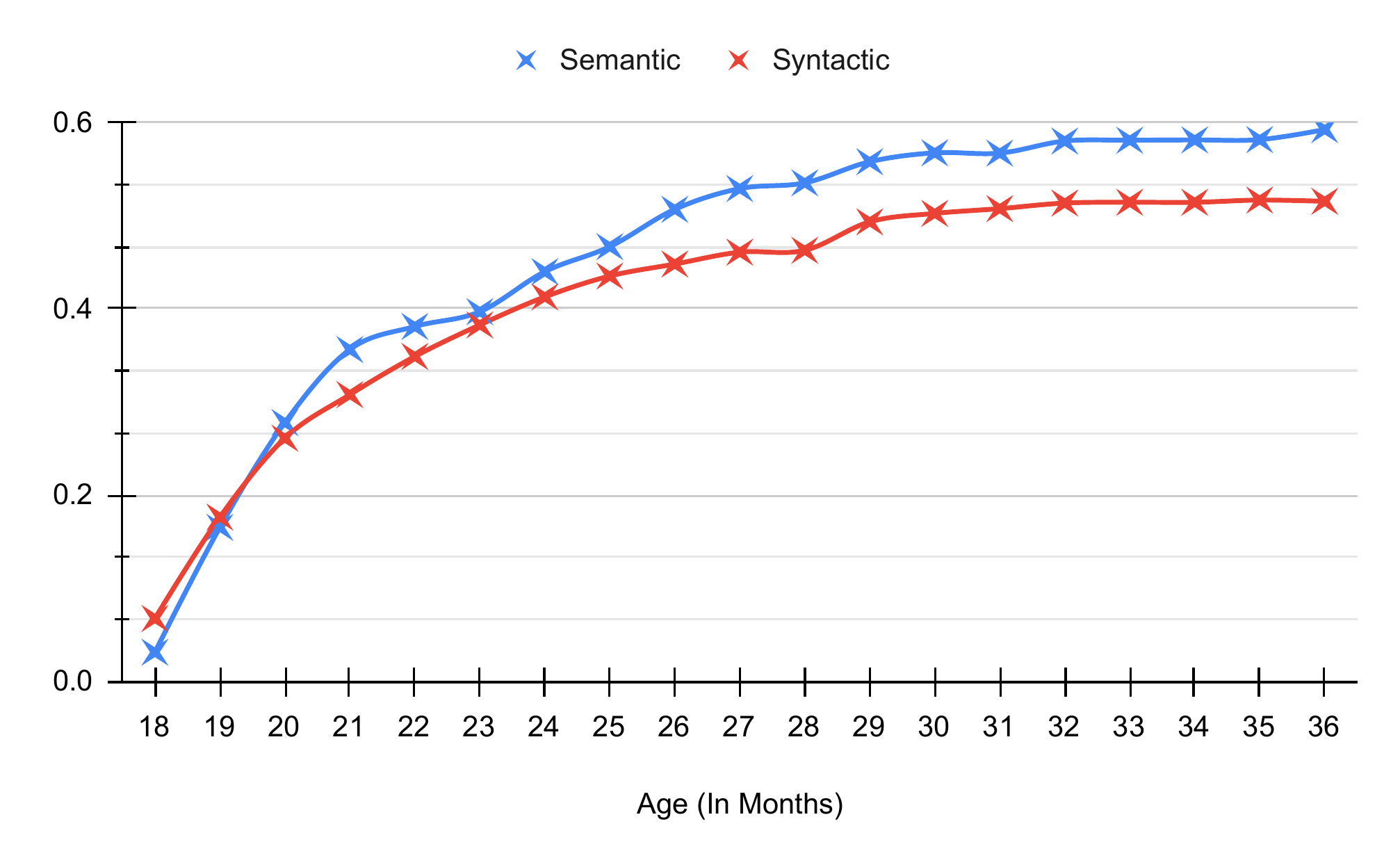}
    \caption{The similarity values between the representations for child-directed speech and child-speech (Spearman's correlation ($\rho$) vs. age in months).}
    \label{fig:rsa}
\end{figure}

The trajectory observed for representational similarities (Figure~\ref{fig:rsa}) across both the semantic and syntactic probe words matches with that of the balanced accuracy from the Lexical Categorization task (Figure~\ref{fig:BA-incremental}). Hence, the incremental diachronic representations successfully capture the fine-grained dynamics of lexical acquisition, which ultimately translates to a higher-level of lexical processing.

\section{Conclusion}
\label{sec-conclusion}
In this paper, we explore the usability of diachronic distributed word representations towards modeling lexical development in children. While all the related previous works use distributed representations with child-directed speech only,\footnote{To the best of our knowledge.} we also obtain the distributed representations of child-speech. This allows us to model the lexical development in children in a more direct way. Through an ablation experiment, we demonstrate the effectiveness of our cognitively motivated incremental learning diachronic model in capturing abstract lexical knowledge in noisy child-speech. We show the usability of our model across various dimensions of the study of lexical development through multiple representative empirical and qualitative analyses. 

Our experiment with the lexical categorization task reveals the trajectories of semantic and syntactic knowledge acquisition in children. Our experiment with the linear mixed-effect modeling of diachronic representational-changes displays the role of input word frequencies in word acquisition. Further, we also perform a fine-grained analysis of lexical knowledge transfer with Representational Similarity Analysis of diachronic representations from child-speech and child-directed adult speech. Our qualitative analyses reveal the phenomena of grounding, abstraction, categorization, and word associations in the mental lexicon of children in an elegant and simple manner (Appendix~\ref{sec:appendix-qualitative}).

\section{Future Directions}
\label{sec-future}
The demonstrated effectiveness and ease of usage of diachronic distributed word representations opens up multiple future directions of research in modeling lexical development. While this work only deals with the usability of our model for word-level studies, the diachronic distributed representations from our model can also be used to study the psycholinguistic development at other granularities as well. Representations for higher granularities (partial-words, syllables, etc.) can be obtained by applying any vector-decomposition method over these word representations. Similarly, the representations for lower granularities (phrasal, clausal, sentence-level, etc.) can also be obtained by applying various vector-pooling techniques over the word representations. Lexical development is usually a multimodal process, where various perceptual modalities are involved. As the data collection efforts for child-speech advance, one can incorporate embeddings from other modalities (phonemic embeddings, visual embeddings, etc.) in modeling the lexical development.

While we use a fairly recent diachronic word embedding model in this work \cite{carlo_bianchi_palmonari}, the lexical modeling efficiency can be increased in parallel with the advances in diachronic word embedding modeling. Further, handling the challenges like partial words, low vocabulary size, lesser training data, etc. can be a good research direction as well. In the future, we plan to extend this work to other languages, using data collected with subjects from a diverse demography.\footnote{The code and data used for this work is available \href{https://github.com/cnrl-bpgc/diachronic-child-lexical-development}{here}.}

\bibliography{anthology,custom}
\bibliographystyle{acl_natbib}

\appendix

\section{Qualitative Analysis}
\label{sec:appendix-qualitative}

\subsection{t-SNE Visualization}
A significant advantage of using distributed representational models is their suitability for qualitative analyses. Given a high-dimensional vector representation space, one can apply various dimensionality reduction algorithms to map it to a two-dimensional space with minimum errors. Which in turn allows one to visualize the representations on a 2D plot. 

\begin{figure*}[htbp]
    \centering
    \includegraphics[width=\textwidth]{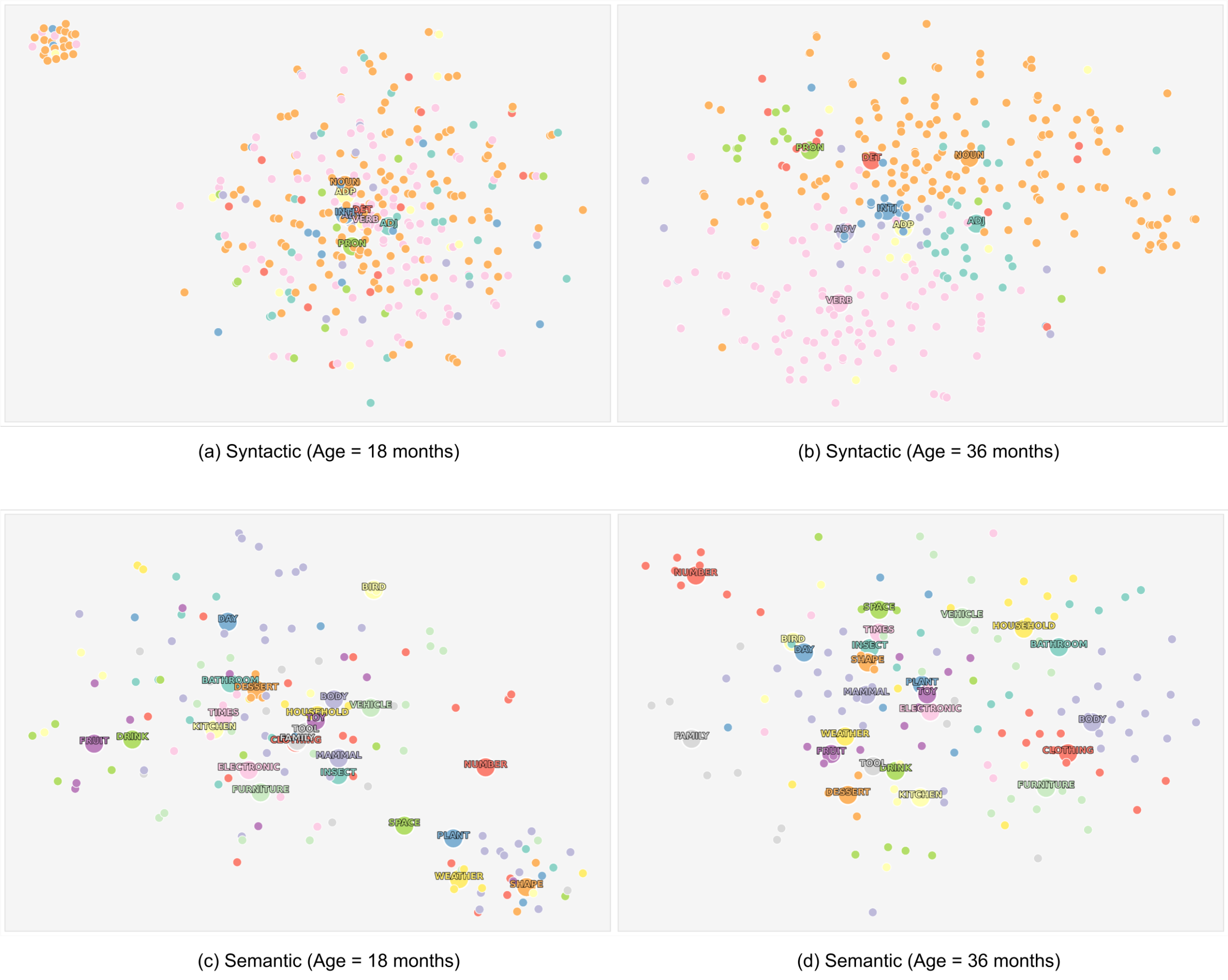}
    \caption{t-SNE visualizations of the probe word categories at Age = $18$ months and Age = $36$ months.}
    \label{fig:tsne}
\end{figure*}

We use a t-SNE dimensionality reduction algorithm \citep{t-sne} on the incremental representations from the child-speech data at Age = $18$ months and Age = $36$ months. We use a perplexity value of $19$ for the t-SNE algorithm. We use the mean vector in each probe-word category to get its $centroid$. To exclude any extreme outliers from the plots, we use Chebyshev's inequality and limit the X-coordinates with a value of $k=8$ standard deviations. We observe several qualitative drifts that emerge in the representations obtained at Age = $36$ months as compared to those obtained at Age = $18$ months (Figure~\ref{fig:tsne}).  

For the semantic categories (Figure~\ref{fig:tsne}d), the words belonging to the categories containing living creatures: $plant$, $mammal$, $insect$, and $bird$ cluster together. The related categories of $household$ and $bathroom$, and the food-related categories: $fruit$, $dessert$ and $drink$ appear together as well. Words in the $clothing$, $body$, and $furniture$ occupy a distinct portion of the space, hinting towards the emergence of grounded word meanings (of clothing) to locations (near furniture like closet and mirror) and usage (over the body). Clusters $day$ and $times$ drift closer as their constituent words are frequently used together. All the food categories appear adjacent to the $kitchen$ category in the space. 

Similar strikingly visible patterns are observed with syntactic categories as well. Unlike the syntactic representations at Age = $18$ months (Figure~\ref{fig:tsne}a), the syntactic representations at Age = $36$ months show well-clustered categories (Figure~\ref{fig:tsne}b). The two major categories: $noun$ and $verb$ become almost linearly separable. Their related categories are sorted accordingly as well. The $pronouns$ and $adjectives$ are placed in the upper half of the plot, occupied by the $noun$ category. On the other hand, $adverbs$ appear in the bottom half of the plot, which is occupied by the $verb$ category. The remaining neutral categories: $determiners$, $interjections$ and $adpositions$ are well placed at the boundary of the $noun$ and $verb$ clusters.

\subsection{Nearest Neighbors}

Another fine-grained approach towards the qualitative analysis of distributed representations is that of observing the nearest neighbors of particular data points.

We note the k-nearest neighbors\footnote{We use cosine-distance to find the nearest neighbors.} $(k=3)$ for a target word from each semantic (Table~\ref{tab:knn-semantic}) and syntactic category (Table~\ref{tab:knn-syntactic}). While the neighbors at Age = $18$ months are a bit random, the neighbors at  Age = $36$ months appear to be more systematically relevant, either by belonging to the same category (example: $zoo \rightarrow \{store, school\}$), or by showing certain abstract free word-associations (example: $tea \rightarrow \{cup, milk\}$;  $wet \rightarrow \{dirty, diaper\}$).

\begin{table*}[htbp]
\centering
\begin{tabular}{|l|l|l|l|}
\toprule
\textbf{Word}       & \textbf{Neighbours at 18 months}                     & \textbf{Neighbours at 36 months}                         & \textbf{Category}   \\ 
 \midrule
 towel      & diaper, floor, neck       & diaper, paper, blanket         & BATHROOM   \\
 duck       & cake, hi, bird            & square, bird, boat             & BIRD       \\
 tummy      & got, your, hurt           & finger, tongue, head           & BODY       \\
 tie        & touch, break, try         & wear, pull, push               & CLOTHING   \\
 today      & maybe, move, many         & camera, kids, bus              & DAY        \\
 cookie     & cookies, still, happy     & cookies, breakfast, strawberry & DESSERT    \\
 tea        & am, heavy, ready          & coffee, cup, milk              & DRINK      \\
 telephone  & talk, doctor, blow        & phone, couch, plate            & ELECTRONIC \\
 mom        & talking, dinner, sleeping & dad, mommy, six                & FAMILY     \\
 strawberry & apple, yep, cheese        & apple, banana, cheese          & FRUIT      \\
 table      & under, chair, sitting     & floor, couch, wall             & FURNITURE  \\
 window     & fell, running, said       & door, kitchen, spider          & HOUSEHOLD  \\
 spider     & fire, wall, moon          & window, sun, bear              & INSECT     \\
 spoon      & floor, hey, side          & fork, bowl, cup                & KITCHEN    \\
 tiger      & eight, blue, green        & dinosaur, chicken, lion        & MAMMAL     \\
 two        & three, four, can          & many, four, five               & NUMBER     \\
 tree       & climb, nap, stand         & wall, climb, square            & PLANT      \\
 square     & ooh, ah, funny            & circle, butterfly, big         & SHAPE      \\
 sun        & pencil, door, wash        & moon, snow, dog                & SPACE      \\
 night      & said, warm, hey           & morning, time, day             & TIMES      \\
 vacuum     & end, running, am          & bike, careful, room            & TOOL       \\
 toy        & pants, running, sweater   & game, block, lion              & TOY        \\
 truck      & fire, man, drive          & tractor, plane, car            & VEHICLE    \\
 snow       & heavy, plane, egg         & sun, wow, grass                & WEATHER \\   
\bottomrule
\end{tabular}
\caption{Nearest semantic neighbours for $k=3$ at the first and last temporal slice.}
\label{tab:knn-semantic}
\end{table*}

\begin{table*}[htbp]
\centering
\begin{tabular}{|l|l|l|l|}
\toprule
\textbf{Word}       & \textbf{Neighbours at 18 months}                     & \textbf{Neighbours at 36 months}                         & \textbf{Category}   \\ \midrule
wet   & water, diaper, baby      & dirty, diaper, hurt  & ADJ  \\ 
with  & different, game, morning & game, lap, help      & ADP  \\ 
where & she, are, how            & yes, who, how        & ADV  \\ 
your  & hands, tummy, feet       & you, yours, okay     & DET  \\ 
yes   & bear, give, blanket      & what, where, yep     & INTJ \\ 
zoo   & bus, mommy, five         & store, school, party & NOUN \\ 
yours & who, touch, noise        & pen, coffee, candy   & PRON \\ 
write & set, ready, touch        & draw, pencil, pen    & VERB \\ 
\bottomrule
\end{tabular}
\caption{Nearest syntactic neighbours for $k=3$ at the first and last temporal slice.}
\label{tab:knn-syntactic}
\end{table*}

\end{document}